\documentclass{article}

\usepackage{graphicx} 
\usepackage{subfigure}
\usepackage{epstopdf}
\usepackage{enumitem}
\usepackage{xfrac}
\usepackage[square,numbers]{natbib}

\usepackage{algorithm}
\usepackage{algorithmic}

\usepackage{amsmath,amsthm}

\usepackage{hyperref}

\newcommand{\suchthat}{\;:\;}
\newcommand{\size}[1]{\left|#1\right|}
\newcommand{\prob}[1]{\operatorname{Pr}\left(#1\right)}
\newcommand{\deter}[1]{\operatorname{det}\left(#1\right)}

\theoremstyle{plain}
\newtheorem{theorem}{Theorem}[section]
\newtheorem{definition}[theorem]{Definition}

\usepackage[accepted]{icml2013}
\icmltitlerunning{Fairness and Diversity}

\begin{document}

\twocolumn[
\icmltitle{ How to be Fair and Diverse?}

\icmlauthor{L. Elisa Celis}{\'{E}cole Polytechnique F\'{e}d\'{e}rale de Lausanne (EPFL), Switzerland}
\icmlauthor{Amit Deshpande}{Microsoft Research, India}
\icmlauthor{Tarun Kathuria}{Microsoft Research, India}
\icmlauthor{Nisheeth K. Vishnoi}{\'{E}cole Polytechnique F\'{e}d\'{e}rale de Lausanne (EPFL), Switzerland}

\vskip 0.2in
]

\begin{abstract}
Due to the recent cases of \emph{algorithmic bias} in data-driven decision-making, machine learning methods are being put under the microscope in order to understand the root cause of these biases and how to correct them.
Here, we consider a basic algorithmic task that is central in machine learning:  subsampling from a large data set.
Subsamples are used both as an end-goal in data summarization (where fairness could either be a legal, political or moral requirement) and to train algorithms (where biases in the samples are often a source of bias in the resulting model).
Consequently, there is a growing effort to modify either the subsampling methods or the algorithms themselves in order to ensure fairness.
However, in doing so, a question that seems to be overlooked is whether it is possible to produce fair subsamples that are also adequately representative of the feature space of the data set -- an important and classic requirement in machine learning. %
Can diversity and fairness be simultaneously ensured?
We start by noting that, in some applications, guaranteeing one does not necessarily guarantee the other, and a new approach is required.
Subsequently, we present an algorithmic framework which allows us to produce both fair and diverse samples.
Our experimental results on an image summarization task show marked improvements in fairness without compromising feature diversity by much, giving us the best of both the worlds.
\end{abstract}

\section{Introduction}
As more and more machine learning algorithms automate data-driven processes in education, recruitment, banking, and judiciary systems, one thing has become evident -- algorithms can have biases \citep{ON2016}.
Given that these algorithms have far-reaching social and economic consequences, it is important to ensure that they comply with non-discrimination and fairness policies based on race, gender, and other sensitive attributes.
Towards this, there is an ongoing effort  to understand and incorporate fairness in machine learning algorithms (e.g., see  \citep{BS2015,IBN2016,Dwork2012,Gummadi2015}).

We study the problem of subsampling a large data set -- a basic task in machine learning.
Subsamples are used both as an end-goal in data summarization (where fairness could either be a legal, political or moral requirement) and to train algorithms (where biases in the samples are often a source of bias in the resulting model).
A crucial requirement for either task is that the sample be {\em diverse} in the feature space;
this is important both to provide a comprehensive viewpoint if the sample is the end-goal, and to make algorithms trained on such samples robust.
Ensuring diversity in samples is well-studied; there are several notions of diversity and approaches for attaining it (see, e.g., \citep{Kulesza2012}).
However, diversity may not guarantee fairness on sensitive attributes and may propagate biases, leading to broken models and algorithmic prejudice \citep{ON2016, BS2015}.
Mathematically, fairness can be viewed as a measure of diversity in the combinatorial space of sensitive attributes, as opposed to the geometric space of features.

\begin{figure}[H]
\centering
\vspace{-.1in}
{\includegraphics[height=5.2cm]{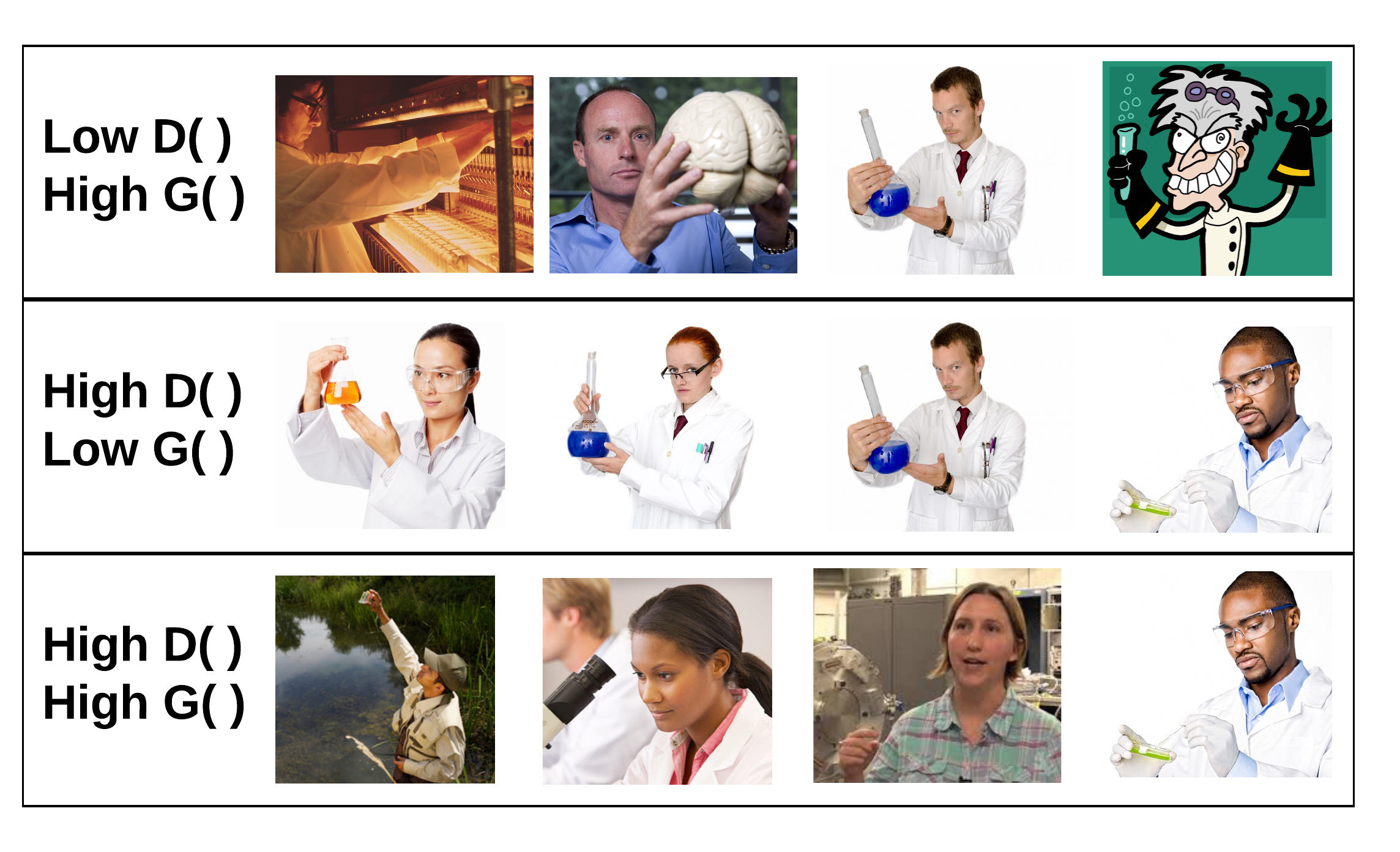}}
\caption{Example sets of images displaying high and low combinatorial and geometric diversity. Our goal is to produce a subset of images that satisfies both notions of diversity; visually distinct and demographically varied,
 as depicted in the bottommost row.}
\label{fig:example}
\end{figure}

This brings us to the central question of this work:
\emph{How do we select samples from a large dataset that are both diverse in features and fair to sensitive attributes?}
Simple examples (such as those in Figure \ref{fig:example}) show that, in certain settings, diversity does not necessarily imply fairness and vice-versa; however, both seem simultaneously achievable.
While both geometric and combinatorial diversities have been studied in independent works  (e.g., \citep{Kulesza2012} and \citep{kamiran2009classifying}), to the best of our knowledge, this is the first systematic study that addresses both \emph{simultaneously}.

\paragraph{Diversity: combinatorial and geometric.}
Formally, we study the following  problem: given a large dataset $X$ of $n$ items, output a geometrically and combinatorially {\em diverse} $S$ of size $k$. To make this well-defined, we need to specify two things: 1) how the data is given and 2) measures of diversity. There are two extremes --
\begin{enumerate}
\item {\bf Combinatorial diversity.} Each data point $x \in X$ has an attribute from a small set $\{1, 2, \dotsc, p\}$, which leads to a combinatorial measure of diversity $D( \cdot )$: The diversity of a set $S$ is the Shannon entropy of the distribution $\left(\sfrac{\size{S_{1}}}{k}, \sfrac{\size{S_{2}}}{k}, \dotsc, \sfrac{\size{S_{p}}}{k}\right)$, where $S_{i}$ is the set of elements in $S$ with attribute value $i$. Intuitively, the larger the entropy, the more diverse is $S$ with respect to the given attributes.
\item {\bf Geometric diversity.} Each data point $x \in X$ has a high-dimensional feature vector $v_{x}$, which motivates a geometric measure of diversity $G( \cdot )$: The diversity of a set $S$ is the (squared) volume of the $k$-dimensional parallelepiped formed by the vectors $\{v_{x} \suchthat x \in S\}$. Intuitively, the larger this volume, the more diverse is $S$ in the feature space.
\end{enumerate}

The combinatorial notion works with much less information, and is known as the \emph{diversity index} \citep{Simpson1949} in social and biological sciences. It is more suited to quantify fairness in sensitive or human-interpretable attributes that take a small set of discrete values.
The geometric notion gives rise to a probability distribution known as \emph{determinantal point process} (or $k$-DPP), and such measures have been used to quantify feature diversity in a variety of machine learning applications for images \citep{Kulesza2012}, videos \citep{GongCGS14}, documents \citep{Lin2012}, recommendation systems \citep{Zhou09}, and sensor placement \citep{Krause2008}.
Besides quantification of diversity in feature-rich datasets, an important reason for the deployment of $k$-DPPs is the recent efficient algorithms to sample from these distributions \citep{DR10,AGR16}.

\paragraph{Our contribution.}
We present an algorithmic framework that allows a user to integrate both notions of diversity, and experimentally demonstrate a marked improvement in fairness without compromising geometric diversity by much -- resulting in the best of both the worlds.

Conceptually, we propose a novel generalization of $k$-DPPs which we call \emph{$P$-DPP}. Given the feature vectors and the partition of the dataset $X = X_{1} \cup X_{2} \cup \dotsb \cup X_{p}$ based on the $p$ different values of a sensitive attribute,  $P$-DPP  samples a $k$-sized subset $S$ with probability proportional to the squared volume of the parallelepiped formed by the feature vectors in $S$ (as is done in $k$-DPPs) but {\em only} over sets $S$ that satisfy ${\size{S \cap X_{i}}}=k_i$, for given $k_i$s for all $1 \leq i \leq p$.
Algorithmically, a polynomial time algorithm for sampling $k$-DPPs generalizes, albeit non-trivially, to $P$-DPPs with a constant number of disjoint partitions (i.e., $p = O(1))$, making our approach feasible \citep{KD16,SV16}.

We experimentally compare the performance of sampling with $P$-DPPs against three natural baselines for an image summarization task.
We consider an image dataset that consists of male and female scientists and artists.
We observe that $P$-DPP outperforms or matches other approaches with respect to both $D(\cdot)$ and $G(\cdot)$ in three different scenarios: 1) when we can ensure perfect fairness, see Section \ref{sec:exp1} and Figure \ref{fig:Exp1}, 2) when some sensitive attributes remain hidden, see Section \ref{sec:exp2} and Figure \ref{fig:Exp2}, and 3) when the underlying dataset is  biased, see Section \ref{sec:exp3} and Figure \ref{fig:Exp3}.

These experiments give strong evidence that sampling with $P$-DPPs is a successful approach for data summarization.
Subsampling is also an important subroutine in various machine learning tasks (see, e.g.,  \citep{DrineasMahoney05,GittensMahoney13}), and it remains an important avenue for future work to study if $P$-DPPs can also help mitigate algorithmic bias in such settings. 

\section{Preliminaries}
Here we give the formal definitions and theoretical constructs used in this paper. An attribute that takes one of $p$ different values gives a natural partition of the underlying data into $p$ parts. The fairness of a dataset (or its subset) with respect to such an attribute can then be quantified by the fairness or diversity index.

\begin{definition}{\bf (Fairness or Diversity Index)}
\label{def:fairness}
Given a set $X$ of $n$ items and its partition $X = X_{1} \cup X_{2} \cup \dotsc \cup X_{p}$ into $p$ parts, the diversity index of any subset $S \subseteq X$ is defined as
the Shannon entropy $
D(S) =- \left(\sum_{i=1}^{p} s_i \log s_i \right)
$
where $s_i = \frac{|S\cap X_i|}{|S|}$.
\end{definition}

For feature-rich data, where a kernel defines the dot product of feature vectors, (sub)determinants extend this notion to define diversity over subsets.
\begin{definition}{\bf (Geometric Diversity)}
\label{def:diversity}
Given a dataset $X$ and a positive semidefinite kernel matrix $K = \left(K(x, y)\right)_{x, y \in X}$, the geometric diversity of a subset $S \subseteq X$ is defined as $G(S) = \deter{K_{S, S}}$, which is the determinant of the principal submatrix $K_{S, S} = \left(K(x, y)\right)_{x, y \in S}$ given by the row and column indices in $S$.
\end{definition}
Geometric diversity defines a distribution on subsets known as a (discrete) determinantal point process.
\begin{definition}{\bf (DPPs and $k$-DPPs)}
\label{def:kdpp}
Given a dataset $X$ and a positive semidefinite kernel matrix $K = \left(K(x, y)\right)_{x, y \in X}$, the DPP is a distribution over subsets $S \subseteq X$ such that the probability $\prob{S} \propto \deter{K_{S, S}}$. The induced probability distribution over $k$-sized subsets is called $k$-DPP.
\end{definition}
Now we define $P$-DPP; our generalization of $k$-DPP to subsets that have the same relative partition as $X$.
\begin{definition}{\bf ($P$-DPP)}
\label{def:pdpp}
Given a dataset $X$, a positive semidefinite kernel matrix $K = \left(K(x, y)\right)_{x, y \in X}$,  a partition $X = X_{1} \cup X_{2} \cup \cdots \cup X_{p}$ into $p$ parts, and numbers $k_1,\ldots,k_p$,  $P$-DPP defines a distribution over $k$-sized subsets $S \subseteq X$ such that $\prob{S} \propto \deter{K_{S, S}}$  if $ |S \cap X_i| = k_i$ and $\prob{S} = 0$, otherwise.
\end{definition}
Lastly, we introduce a natural baseline which we also compare against in our experiments.
\begin{definition}{\bf ($k_i$-DPP)}
\label{def:kidpp}
Given a dataset $X$, a positive semidefinite kernel matrix $K = \left(K(x, y)\right)_{x, y \in X}$,  a partition $X = X_{1} \cup X_{2} \cup \cdots \cup X_{p}$ into $p$ parts, and numbers $k_1,\ldots,k_p$,  $k_i$-DPP defines a distribution over $k_1+k_2+\cdots+k_p$-sized subsets $S \subseteq X$ that is a product distribution:  for each $i,$ we obtain a  sample $S_{i} \subseteq X_{i}$ of size $k_i$ independently with probability proportional to $\prob{S_i} \propto \deter{K_{S_i, S_i}}$, and combine these samples to output $S = S_{1} \cup S_{2} \cup \dotsb \cup S_{p}$.
\end{definition}
We emphasize  that the  difference between a $k_i$-DPP and  a $P$-DPP with the same parameters $(k_1,\ldots,k_p)$ is that the samples $S_i$ from each part in $P$-DPP are not {\em independent} as in $k_i$-DPP. Indeed, this is what makes them more powerful.

Polynomial time sampling from $k$-DPPs uses a linear algebraic fact that the partition function as well as the marginals of $k$-DPP can be computed using the characteristic polynomial of the underlying kernel matrix. A multivariate generalization of this can incorporate partition constraints (and beyond) to sample from $P$-DPPs in time $n^{O(p)}$, which is polynomial for $p = O(1)$ \citep{KD16,SV16}.

\section{Experimental Results}
\renewcommand{\thesubfigure}{(\roman{subfigure})}

\subsection{Datasets and Features}

We ran our experiments on a collection of images curated using Google image search as follows:
Four search terms were used: (a) ``Scientist Male'', (b) ``Scientist Female'', (c) ``Painter Male'', and (d) ``Painter Female''. The search was restricted to medium sized JPEG files that passed the strictest level of Safe Search filtering. The top 200 distinct images from each were collected to create the following three datasets:\footnote{The images are available at \url{goo.gl/hNukfP}.}
\begin{itemize}[noitemsep, topsep=0pt]
\item Scientist: (a) and (b)
\item Artist: (c) and (d)
\item Scientist+Artist: (a), (b), (c), and (d).
\end{itemize}
Hence, each dataset has inherent labels (a)-(d) over which we can measure the combinatorial diversity of a sample. In order to measure geometric diversity, following \citep{Kulesza2011}, each image was processed with the \texttt{vlfeat} toolbox to obtain sets of 128-dimensional SIFT descriptors \citep{Lowe99,vlfeat}. The descriptors are combined, subsampled to a set of 36,000 and then clustered using $k$-means into 256 clusters. The feature vector for an image is the normalized histogram of the nearest clusters to the descriptors in the image. Finally, the kernel value $K(x,y)$ for any pair of images $x$ and $y$ is obtained by taking the dot-product of the SIFT features of $x$ and $y$.

\subsection{Algorithms and Baselines}

In each experiment, we compare four different probability distributions from which to select $k$ samples from a dataset: 1) Our proposed $P$-DPP (see Def~\ref{def:pdpp}), 2) the classic $k$-DPP (see Def~\ref{def:kdpp}), 3)  $k_i$-DPP (see Def~\ref{def:kidpp}), and  4) UNIF, which takes a uniformly random subset of size $k$.

In order to sample from $k$-DPP, $k_i$-DPP and $P$-DPP, instead of using the polynomial time algorithms of \citep{KD16,SV16}, we appeal to a Markov Chain Monte Carlo (MCMC) heuristic  inspired by \citep{AGR16} as the latter seems faster in practice.
The Markov chain is defined over the space of subsets of cardinality $k$. The algorithm first chooses a ``warm start state'' $S$ obtained by greedily maximizing the determinant while satisfying the partition constraints. Then, in each iteration, elements $i \in S$ and $j \not\in S$ are chosen uniformly at random. The chain moves to state $T = S \setminus \{i\} \cup \{j\}$ with probability $\frac{1}{2}\min\{1, \sfrac{\det(K_{T,T})}{\det(K_{S,S})}\}$, if it satisfies the constraints. Otherwise, it stays in state $S$. This is repeated for a suitable number of iterations  to guarantee that samples drawn from this chain are ``close'' to that of the desired distribution. %
In each experiment, given a sample $X_{\mathcal A}$ selected by algorithm $\mathcal A$, we report the combinatorial diversity using the fairness index $D(X_{\mathcal A})$ (see Def~\ref{def:fairness}) 
and the geometric diversity $G(X_{\mathcal A})$ (see Def~\ref{def:diversity}).

\subsection{Experiments and Discussion}


\subsubsection{Experiment 1: Perfect Fairness}
\label{sec:exp1}

\begin{figure*}
\hspace{-.25in}\subfigure{\includegraphics[height=3cm]{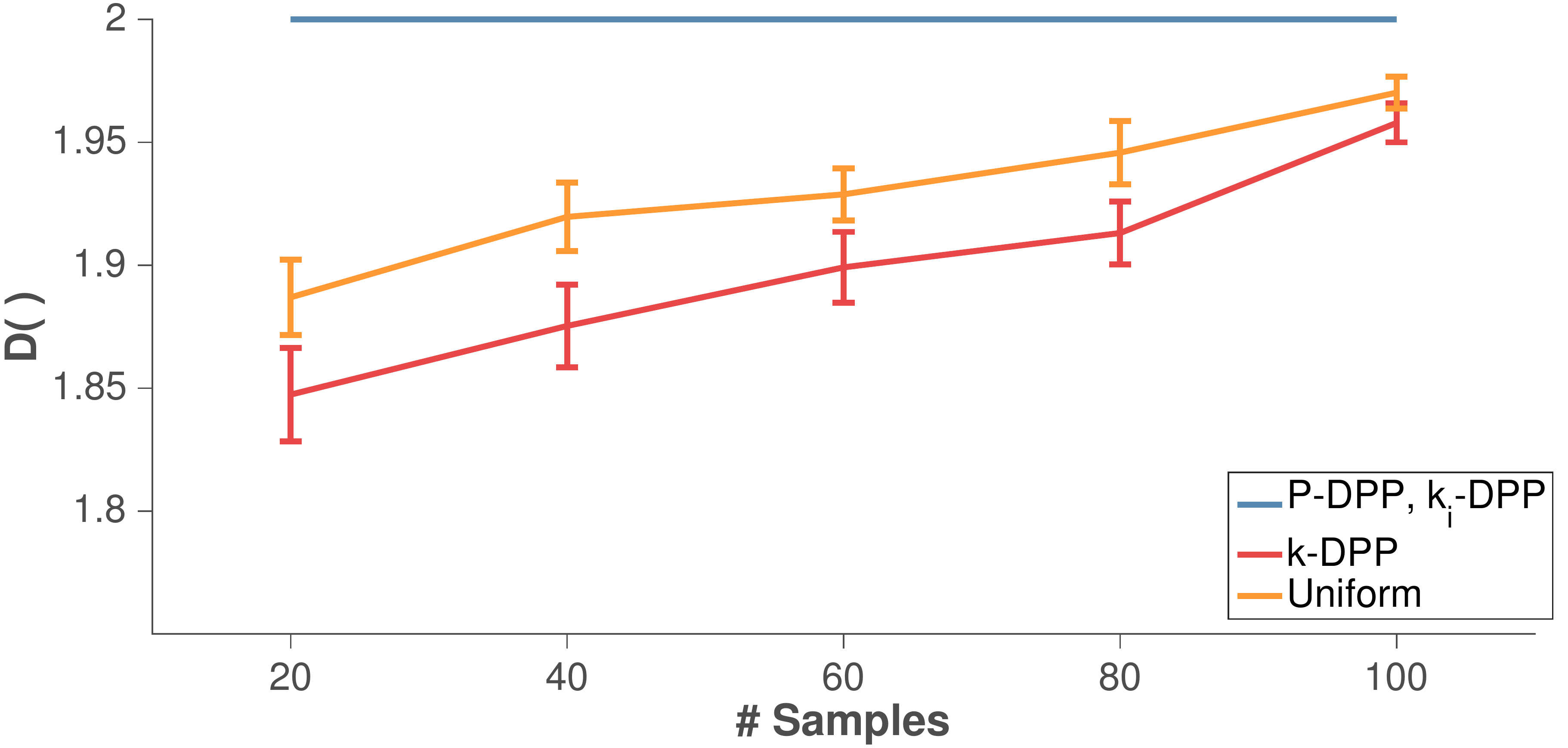}}
\;\;\subfigure{\includegraphics[height=3cm]{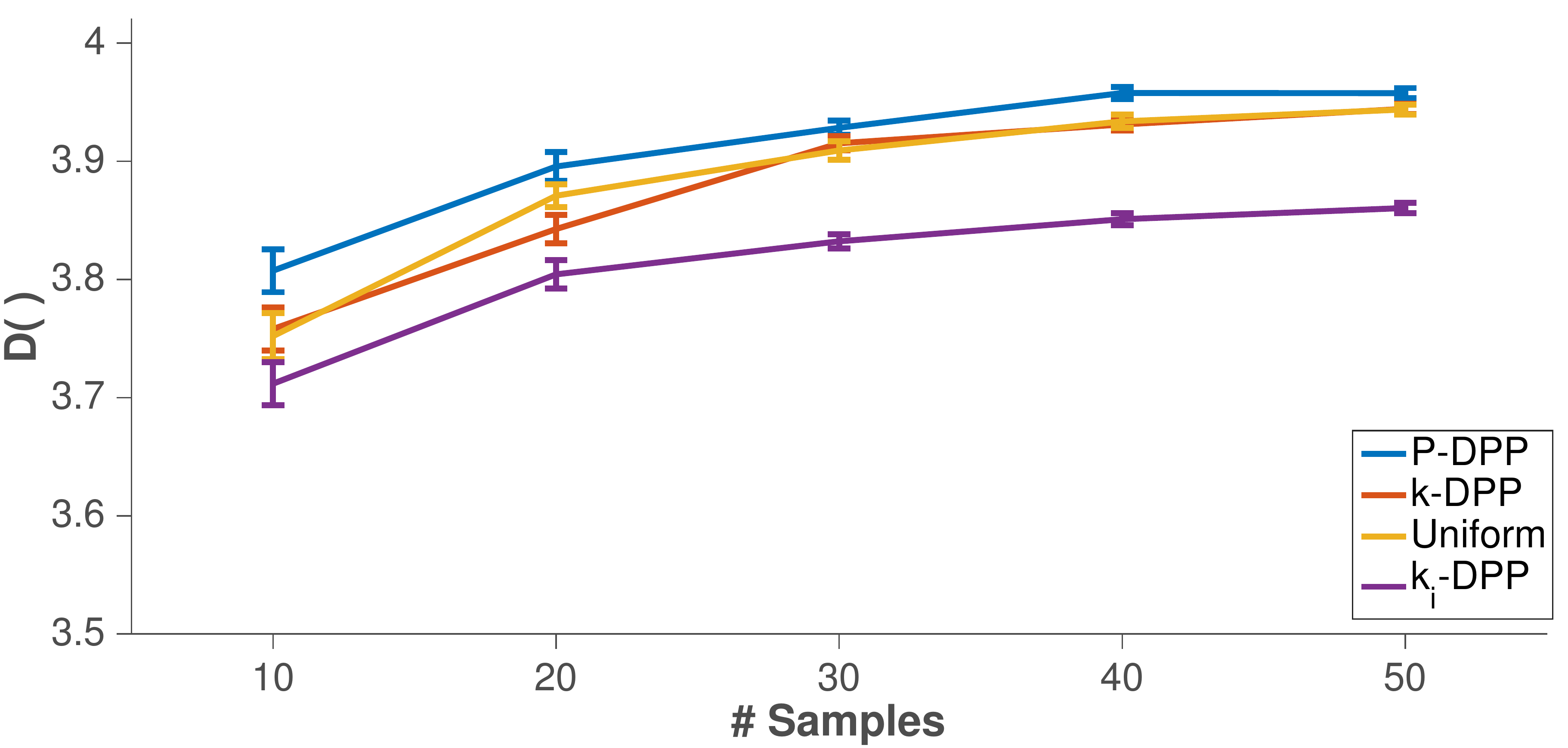}}
\;\;\subfigure{\includegraphics[height=3cm]{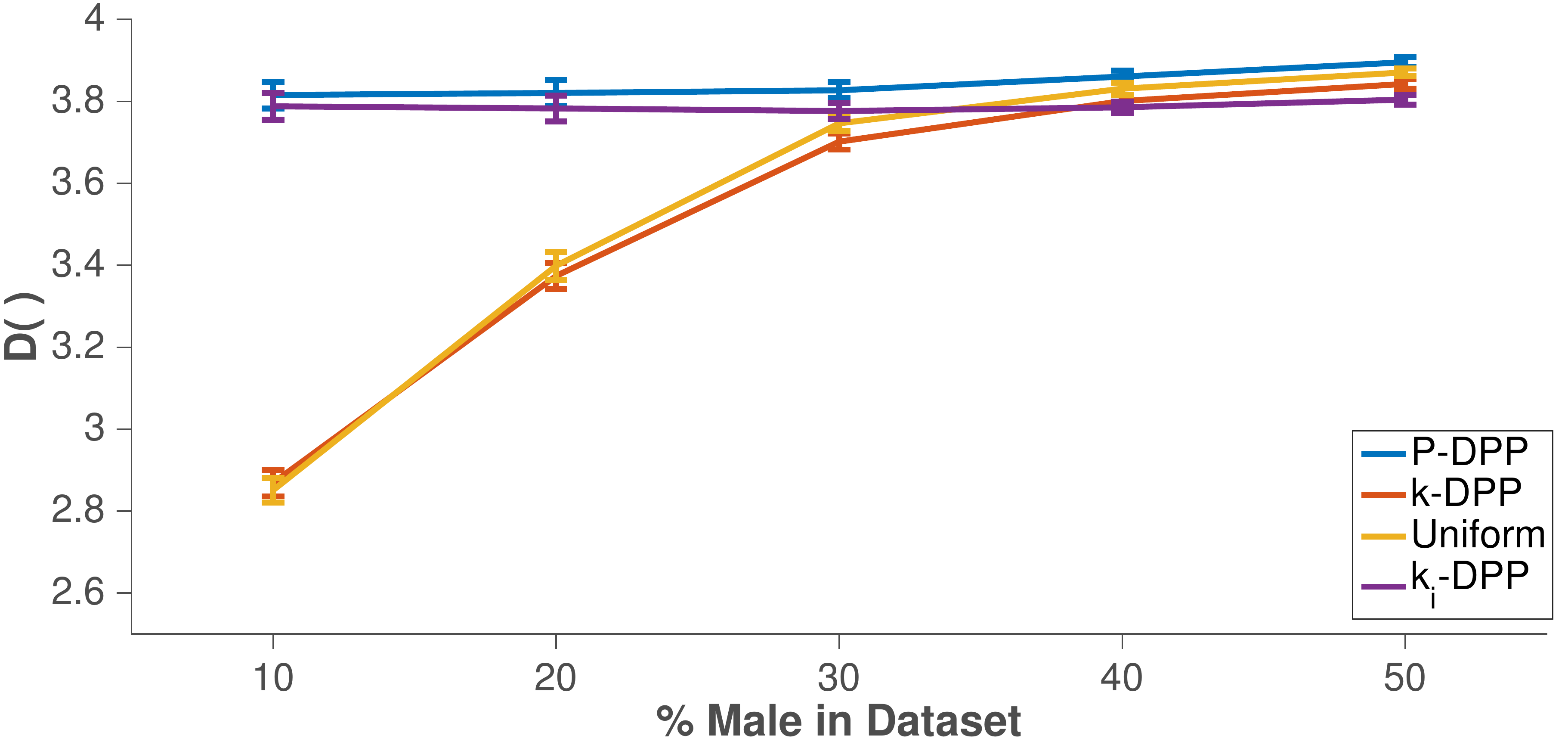}}

\setcounter{subfigure}{0}
\hspace{-.25in}\subfigure[Exp. 1: Performance on the Scientist dataset as the number of samples $k$ increases.
\label{fig:Exp1}]
{\includegraphics[height=3cm]{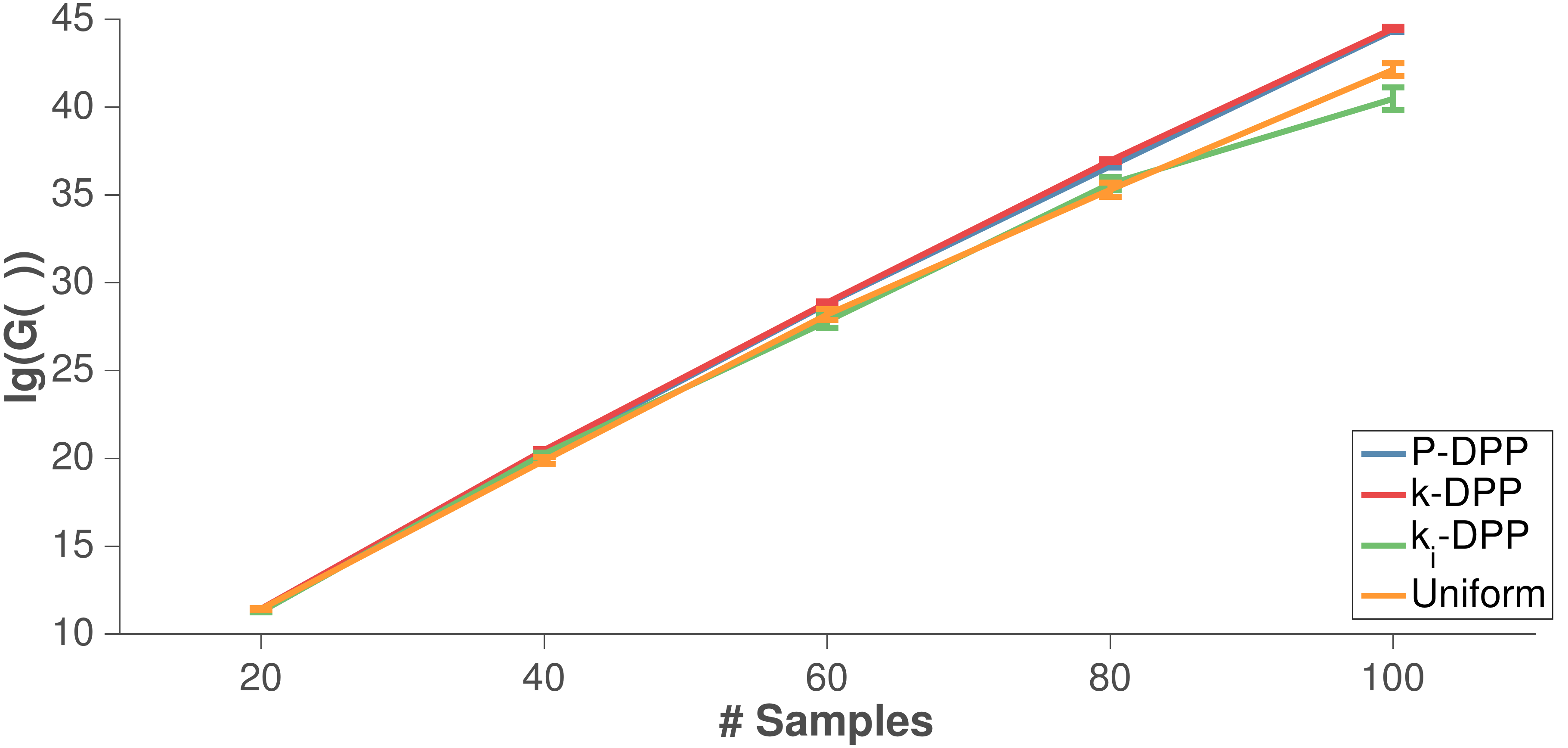}}
\;\;\subfigure[Exp. 2: Performance on the Scientist+Artist dataset as the number of samples $k$ increases. 
\label{fig:Exp2}]
{\includegraphics[height=3cm]{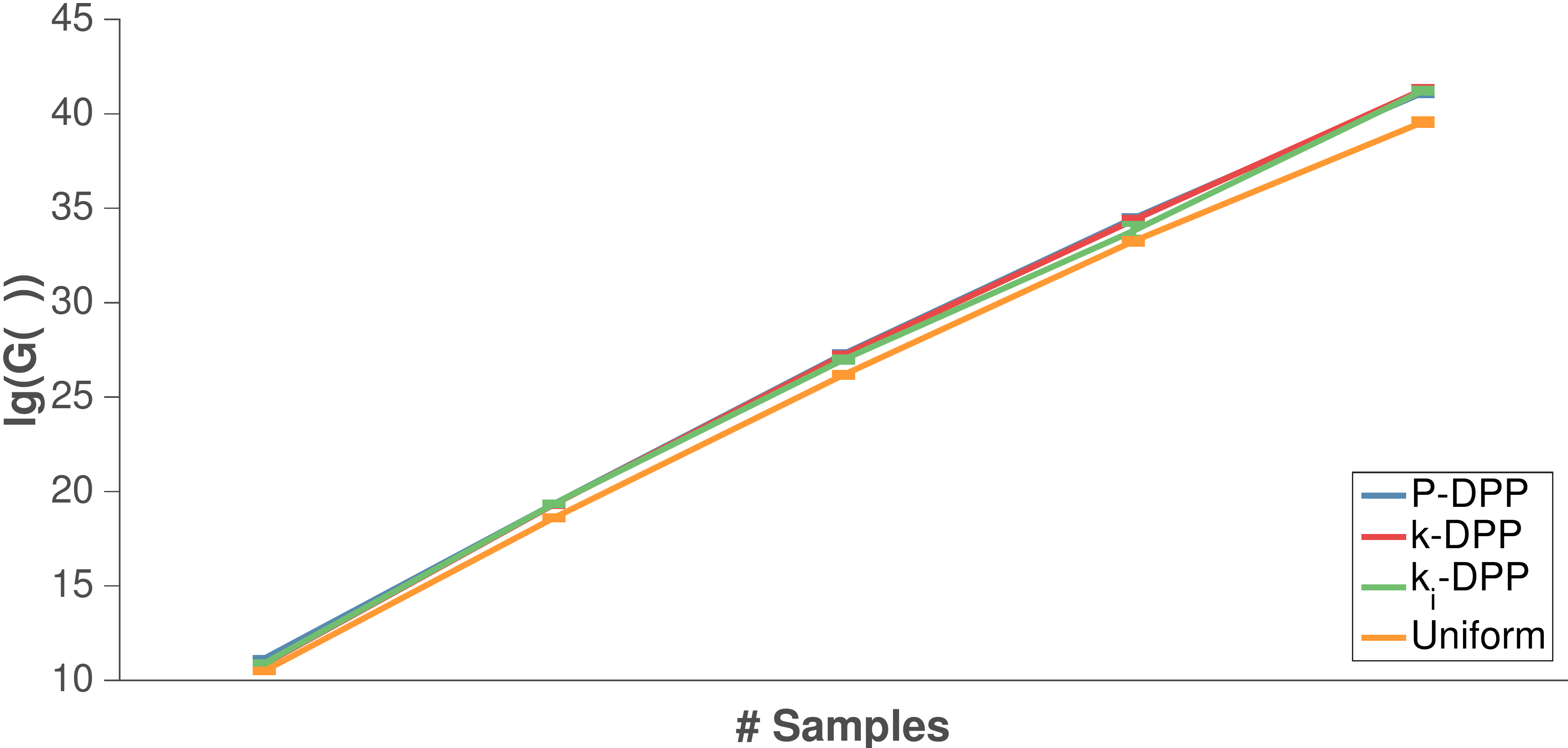}}
\;\;\subfigure[Exp. 3: Performance on the Scientist+Artist dataset for $k=40$ as the bias in the underlying dataset decreases.
\label{fig:Exp3}]
{\includegraphics[height=3cm]{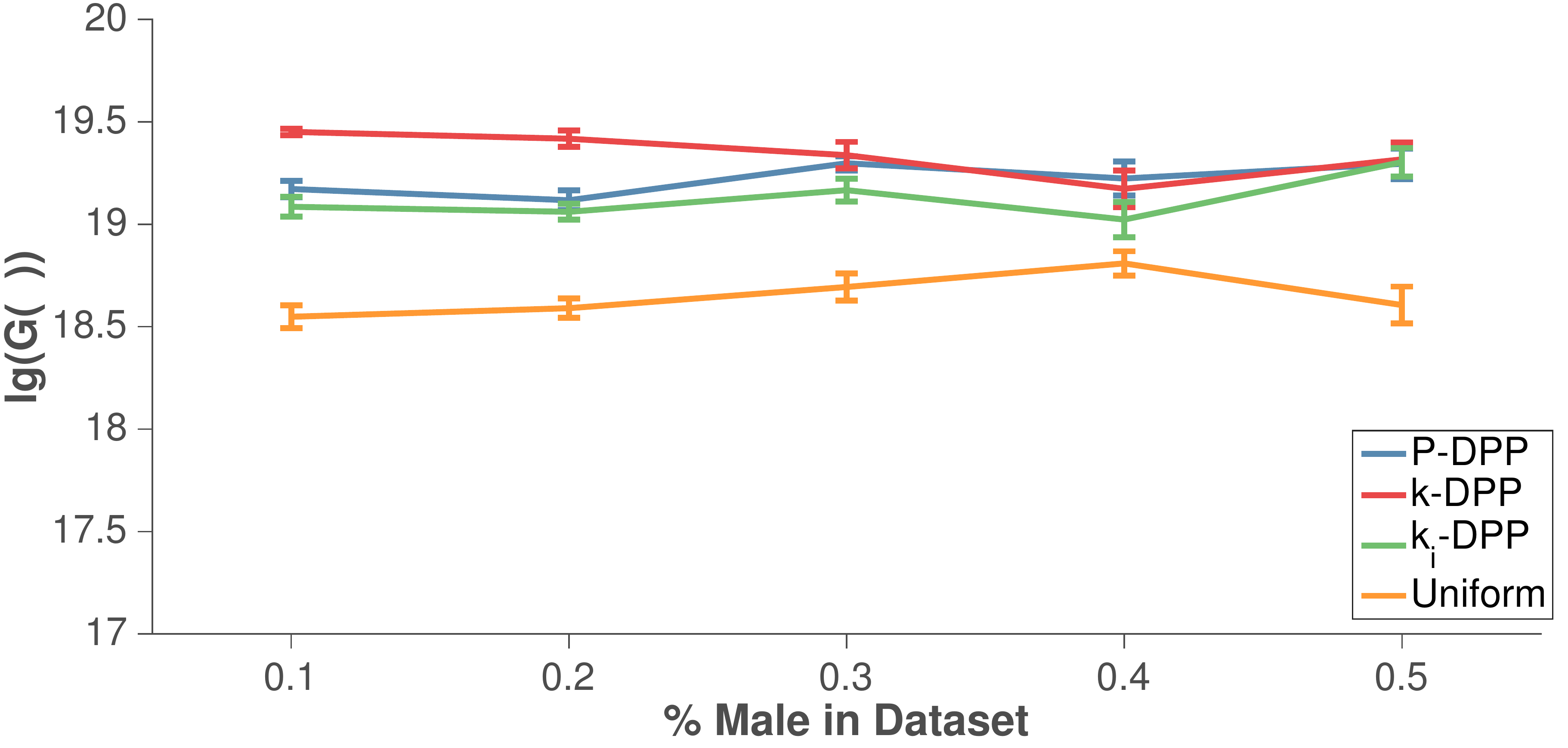}}
\caption{For each experiment the mean fairness index $D(\cdot)$ and log of the geometric diversity $ln(G(\cdot))$ are reported in the top and bottom figures respectively for $n = 100$ repetitions. Error bars represent the standard error of the mean.}
\label{fig:experiments}
\end{figure*}

\paragraph{Experimental Setup.}
We first consider the performance as we vary the sample size $k$ from 20 to 100 on the Scientist dataset (see Figure~\ref{fig:Exp1}); recall that the dataset has two parts, male and female, and that the dataset is unbiased. We place fairness constraints so that $P$-DPP and $k_i$-DPP select exactly 50\% of their samples from the male and female parts. Hence, we have set up the experiment to guarantee optimal $D(\cdot)$ for  $P$-DPP and $k_i$-DPP, and measure the resulting degradation in $G(\cdot)$.

\paragraph{Results.}
Both $P$-DPP and $k_i$-DPP attain the optimal $D(\cdot)$ of $2$. As expected, this is significantly higher than UNIF and $k$-DPP (paired one-sided t-tests, $p < 0.05$). In fact, even UNIF has significantly higher fairness than $k$-DPP (paired one-sided t-test, $p < 0.05$). 
With respect to $G(\cdot)$, the performance of $k$-DPP and $P$-DPP is comparable, with neither significantly outperforming the other. This is notable as $P$-DPP has constraints that $k$-DPP need not abide by; hence, a priori, $k$-DPP could be significantly better. Moreover, both $k$-DPP and $P$-DPP have significantly higher $G(\cdot)$ than UNIF and $k_i$-DPP (paired one-sided t-tests, $p < 0.05$). Outperforming UNIF is expected as random selection makes no effort to increase $G(\cdot)$, however the outperformance of $k_i$-DPP is notable for two reasons: 1) $k_i$-DPP is the only other algorithm that matched the fairness index of $P$-DPP, and 2) $k_i$-DPP is also explicitly attempting to improve $G(\cdot)$. However, while $k_i$-DPP improves $G(\cdot)$ \emph{within} a part of the dataset, it does not diversify \emph{across} parts; %
$P$-DPP avoids exactly this pitfall.

\paragraph{Conclusion.}
This experiment demonstrates that $P$-DPP can match or outperform the other approaches with respect to both fairness and diversity. This conclusion is not unique to this dataset -- we also conducted the same experiment on the Artist dataset, and the results are very similar with the same significance findings holding; we omit the full details due to length constraints. Exploring whether such results are consistent on other types of datasets would be a clear direction for future work.

\subsubsection{Experiment 2: Hidden Attributes}
\label{sec:exp2}

\paragraph{Experimental Setup.}
We then consider the performance of the algorithms as we vary the sample size $k$ from 10 to 50 on the Scientist+Artist dataset, but consider the case where there is a hidden underlying partition (see Figure~\ref{fig:Exp2}). Here, we place fairness constraints so that $P$-DPP and $k_i$-DPP select exactly $50\%$ of their samples from the male (a and c) images and female (b and d) images, but \emph{do not} enforce constraints across scientist (a and b) images and artist (c and d) images, allowing for disproportionality across this dimension. However, we measure the fairness with respect to all four parts.

\paragraph{Results.}
With respect to the $D(\cdot)$, $P$-DPP and $k_i$-DPP no longer attain the optimal fairness of 4. However, $P$-DPP significantly outperforms $k$-DPP, UNIF \emph{and} $k_i$-DPP (paired one-sided t-tests, $p < 0.05$), with $k_i$-DPP being the worst performer despite the partial constraints.
With respect to $G(\cdot)$, as in Experiment 1, the performance of $k$-DPP and $P$-DPP is comparable, and both have significantly higher $G(\cdot)$ than UNIF (paired one-sided t-tests, $p < 0.05$). For this experiment, $P$-DPP is also comparable to $k_i$-DPP, with a mean determinant that is higher, but not significantly so; this is largely due to the fact that for this experiment $k$ is smaller while the dataset size is larger, and hence the drop-off in performance of $k_i$-DPP is not as evident as it was in Experiment 1.

\paragraph{Conclusion.}
Hence, this experiment demonstrates that $P$-DPP can match or outperform the other approaches with respect to both fairness and diversity, even when some of the underlying attributes are unknown. This is an important consideration as we should not inadvertently boost one kind of fairness at the expense of another.

\subsubsection{Experiment 3: Biased Datasets}
\label{sec:exp3}

\paragraph{Experimental setup.}
Lastly, we consider the situation where the underlying dataset is biased (see Figure~\ref{fig:Exp3}). We include all female (b and d) images, but only include a subsample of male images (a and c) in the dataset in order to create biased datasets that have between $10\%$ to $50\%$ male images. The subsampled images are selected uniformly at random from all male scientists and artists for each repetition in the experiment.
We place fairness constraints so that $P$-DPP and $k_i$-DPP select exactly $50\%$ of their samples from the male (a and c) images and female (b and d) images, \emph{regardless of the bias in the underlying dataset}. As in Experiment 2, we \emph{do not} enforce constraints across scientist (a and b) images and artist (c and d) images, but measure $D(\cdot)$ with respect to all four attributes.

\paragraph{Results.}
With respect to $D(\cdot)$, $P$-DPP significantly outperforms $k$-DPP, UNIF and $k_i$-DPP (paired one-sided t-tests, $p < 0.05$). Here, we see that the bias in the underlying dataset can dramatically affect the fairness of UNIF and $k$-DPP as neither approach is designed to correct for such biases. However, $P$-DPP and $k_i$-DPP are able to remain relatively stable throughout.
With respect to $G(\cdot)$, $P$-DPP has significantly higher $G(\cdot)$ than UNIF and $k_i$-DPP (paired one-sided t-tests, $p < 0.05$). However, now $k$-DPP significantly outperforms $P$-DPP (paired one-sided t-test, $p < 0.05$). This is due to the fact that when the dataset is highly biased, the available selection of images in the smaller partition is limited, and hence it is more difficult for $P$-DPP to diversify across the feature space. Indeed, we expect this gap to close as the size (but not proportion) of the smaller part increases.

\paragraph{Conclusion.}
In this experiment we observe that, when the underlying data is highly biased, there is now a tradeoff between $D(\cdot)$ (for which $P$-DPP performs best) and $G(\cdot)$ (for which $k$-DPP performs best). Despite these differences, we note that the gap in $P$-DPP's geometric diversity gradually decreases, while $k$-DPPs fairness index drops rapidly as the bias increases, leading us to conclude that $P$-DPPs remain the best of both worlds, allowing for fairness \emph{and} diversity. 

\small{
\bibliographystyle{plain}
\bibliography{references}
}

\end{document}